\documentclass[journal]{IEEEtran}

\usepackage[utf8]{inputenc}
\usepackage[numbers,sort&compress]{natbib}

\usepackage[table]{xcolor}
\usepackage{times}
\usepackage{epsfig}
\usepackage{graphicx}
\usepackage{rotating}
\usepackage{amsmath}
\usepackage{amssymb}
\usepackage{booktabs}
\usepackage{float}
\usepackage{placeins}
\usepackage{multirow}
\usepackage{balance}
\usepackage{multirow}
\usepackage{subfigure}
\usepackage[flushleft]{threeparttable}
\usepackage{textcomp}
\usepackage{url}

\begin{document}
    
        \title{Towards Automated Melanoma Screening: \\ Proper Computer Vision \& Reliable Results}
    

    \author{
        Michel Fornaciali, 
        Micael Carvalho, 
        Flávia Vasques Bittencourt, 
        Sandra Avila, 
        Eduardo Valle
    \thanks{M. Fornaciali, M. Carvalho, S. Avila and E. Valle are with the School of Electrical and Computing Engineering, University of Campinas (UNICAMP), Campinas, Brazil (e-mail \{michel, micael, sandra, dovalle\}@dca.fee.unicamp.br).}
    \thanks{F. V. Bittencourt is with the School of Medicine, Federal University of Minas Gerais (UFMG), Belo Horizonte, Brazil.}
    }
    

    \maketitle
    
    \begin{abstract}
        In this paper we survey, analyze and criticize current art on automated melanoma screening, reimplementing a baseline technique, and proposing two novel ones. Melanoma, although highly curable when detected early, ends as one of the most dangerous types of cancer, due to delayed diagnosis and treatment. Its incidence is soaring, much faster than the number of trained professionals able to diagnose it. Automated screening appears as an alternative to make the most of those professionals, focusing their time on the patients at risk while safely discharging the other patients. However, the potential of automated melanoma diagnosis is currently unfulfilled, due to the emphasis of current literature on outdated computer vision models. Even more problematic is the irreproducibility of current art. We show how streamlined pipelines based upon current Computer Vision outperform conventional models --- a model based on an advanced bags of words reaches an AUC of 84.6\%, and a model based on deep neural networks reaches 89.3\%, while the baseline (a classical bag of words) stays at 81.2\%. We also initiate a dialog to improve reproducibility in our community.
    \end{abstract}
    

    %

    \section{Introduction}

In this paper we survey recent techniques for automated melanoma screening, broadly and in-depth, discussing the challenges to advance the research. To that end, we reimplement and analyze in detail one existing technique, and contrast it to two novel propositions inspired on modern Computer Vision.

Melanoma is among the most curable cancers when detected early, but ends as one of the most lethal due to delays in diagnosis and treatment, killing one person every hour. The prognosis quickly worsens as the disease progresses, due to its tendency to rapidly metastasize.

\begin{figure}[tb]
    \centering
    \subfigure{\includegraphics[width=0.23\columnwidth]{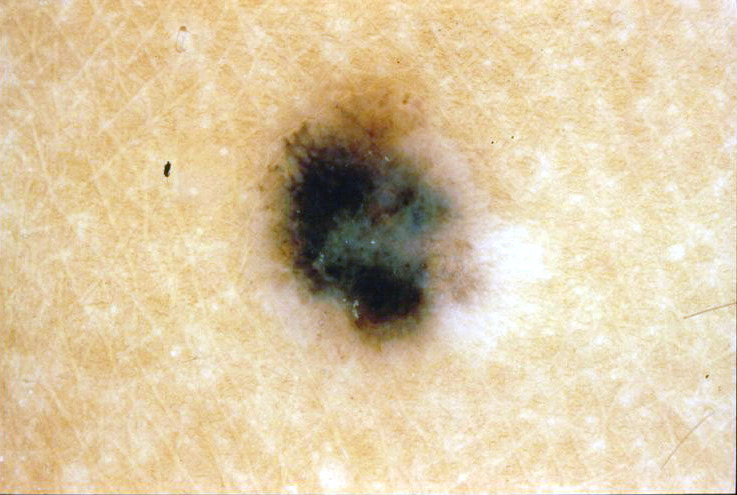}} \subfigure{\includegraphics[width=0.23\columnwidth]{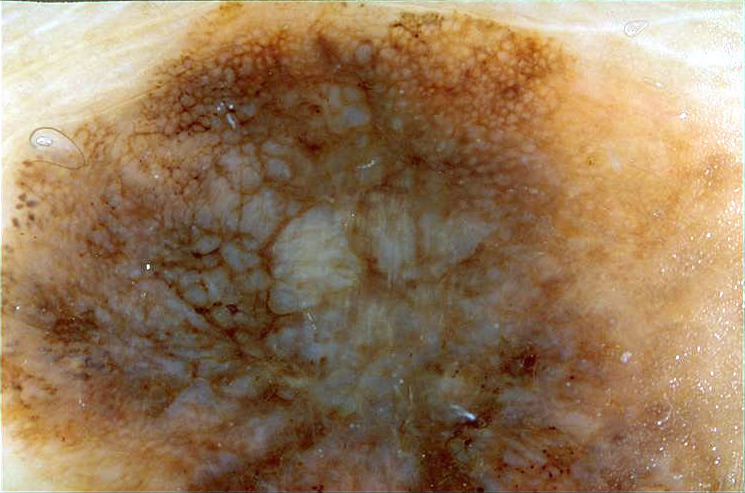}}
    \subfigure{\includegraphics[width=0.23\columnwidth]{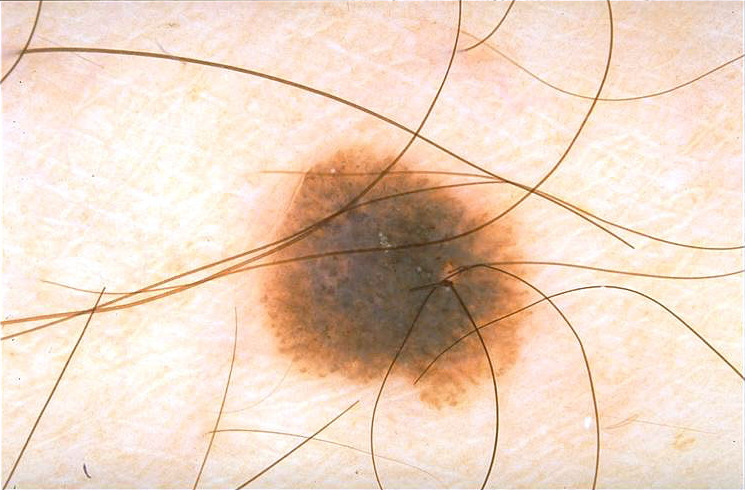}} 
    \subfigure{\includegraphics[width=0.23\columnwidth]{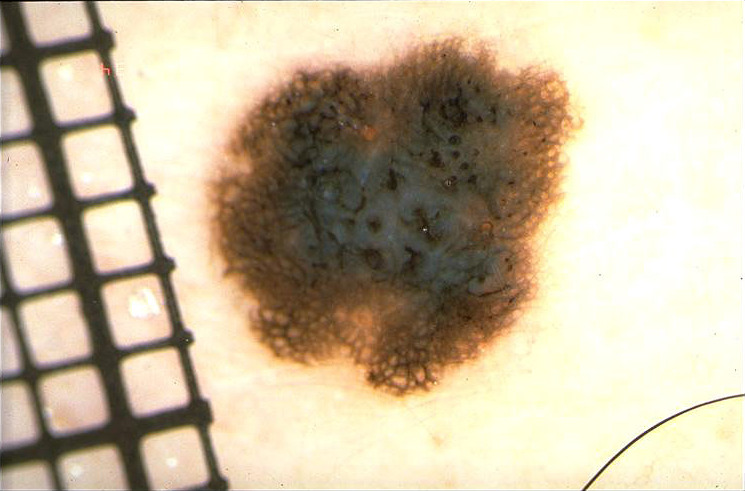}}
    \subfigure{\includegraphics[width=0.23\columnwidth]{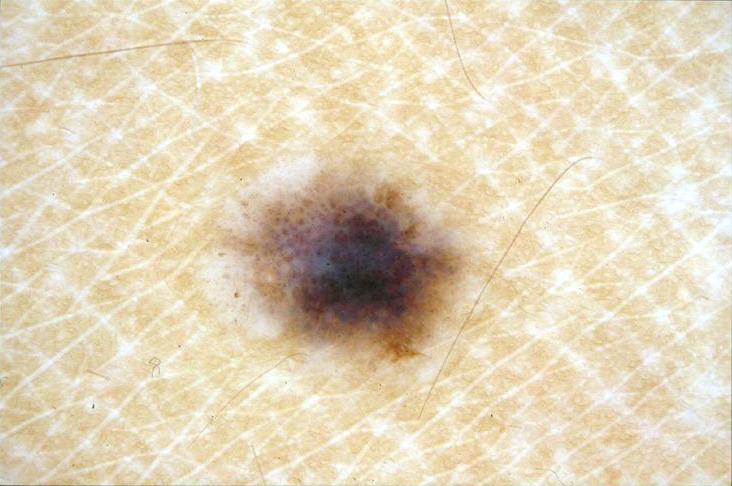}}
    \subfigure{\includegraphics[width=0.23\columnwidth]{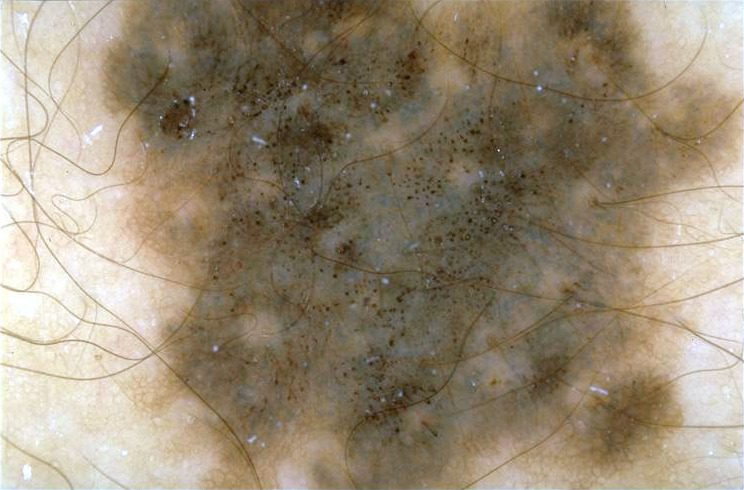}} 
    \subfigure{\includegraphics[width=0.23\columnwidth]{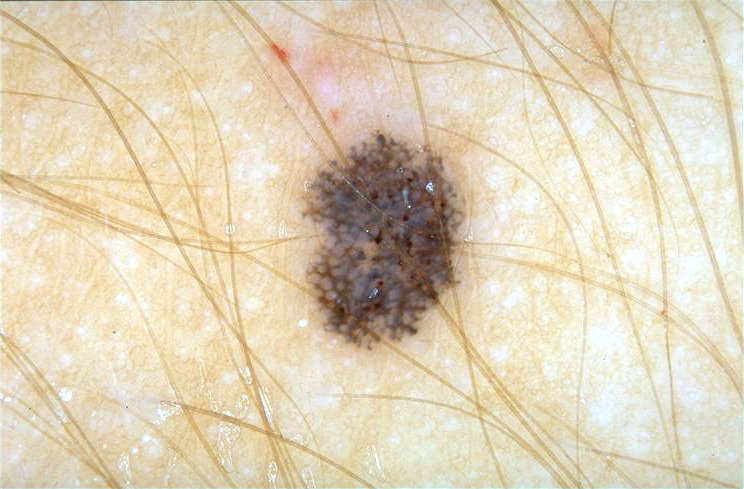}}
    \subfigure{\includegraphics[width=0.23\columnwidth]{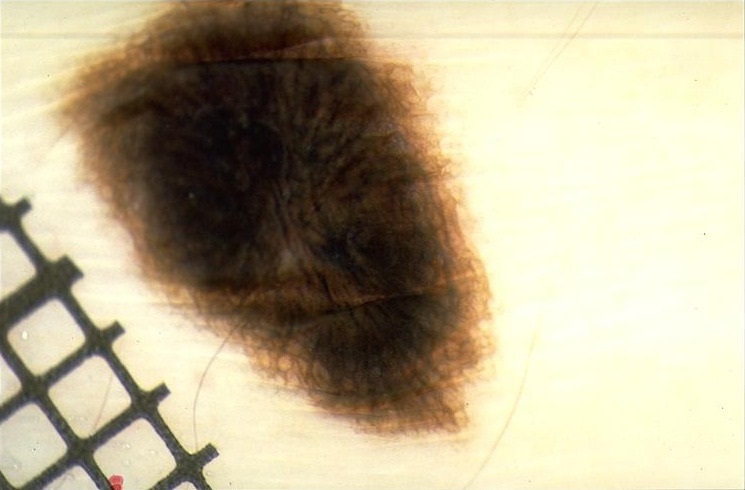}}
	\caption{Extracts of skin lesions from the dataset employed in this work (Section~\ref{sec:dataset}). Melanomas (top row) looks similar to other skin lesions (bottom row), hindering diagnosis both for humans and machines. 
	}
	\label{fig:dataset}
\end{figure}

Melanoma incidence is soaring. In the 1930s, 1 in 1\thinspace500 USA residents developed the disease; in the 2010s that incidence jumped to 1 in 59~\citep{rigel2010epidemiology}. Medical systems, public and private, are struggling to keep up with those rates, since the number of specialists able to diagnose the disease has not grown proportionally. Melanoma is difficult to diagnose reliably (Figure~\ref{fig:dataset}), requiring extensively trained specialists. Improving the diagnosis of melanoma is an urgent need. 

Automated screening appears as an alternative to alleviate the problem~\citep{iyatomi08Computerized, situ08EMBC, wadhawan11EMBC, wadhawan11ISBI, abbas12Skin, doukas12EMBC, marques12EMBC, mikos12BIOMEP, barata13iciar, abuzaghleh2014automated, barata2014two, barata2015improving, abedini2015generalized}. Those techniques are still too inaccurate to give definitive diagnosis, but may soon become accurate enough to assist medical personnel in the triage of patients. Focusing specialists' time on the patients at risk, while safely discharging the others, can greatly increase the quality and cost-effectiveness of care. Such triage can greatly benefit (rural, isolated, or poor) communities that cannot keep a specialist full time; and also help any primary care provider in the difficult decision of referring or not each patient to specialists.

However, current art on automated screening fails to harness the full potential of Computer Vision, by employing outdated models. More worrisome: current art is essentially irreproducible, due to insufficient description of methods, compounded by the widespread practice of withholding code and data. Although we do not presume to exhaust that issue in this work, we invite researchers interested on automated melanoma screening to start a dialogue for improving the situation. In that spirit, we strive to make this paper reproducible, and discuss some guidelines at the final section. 

We list below our main contributions:
\begin{itemize}
    \item we propose a melanoma screening pipeline with two modern Computer Vision techniques, and show that they outperform a classical pipeline;
    \item we show that automated screening pipeline can be streamlined, forgoing hair elimination and lesion segmentation;
    \item we survey the recent literature, providing a broad, in-depth, critical analysis;
    \item we dissect the best-described technique from literature, showing that current art is effectively irreproducible;
    \item we discuss guidelines to improve reproducibility and experimental validity in our community.
\end{itemize}

A preliminary work of ours \citep{Fornaciali2014} explored advanced BoVW for melanoma screening. That work was  focused on fine-tuning the BoVW pipeline for melanoma screening, and was validated in a  much smaller dataset than the current paper. Although we were already worried about reproducibility in that preliminary work, the current paper is much  more mature in that effort.

A short organization summary: in Section~\ref{sec:soa} we survey the current art on melanoma screening from the viewpoint of Computer Vision, while in Section~\ref{sec:problems} we dissect its main problems (on Section~\ref{sec:baseline} we discuss in-depth our attempt to reproduce a particular work, which we use as baseline). In Section~\ref{sec:proposed} we introduce two novel solutions. In Section~\ref{sec:experimental} we propose a validation procedure, whose results, with analyses, appears in Section~\ref{sec:results}. In Section~\ref{sec:conclusions} we discuss our findings, and resume the discussion on reproducibility and experimental validity, with tentative guidelines.

    \section{Review of Existing Art} 
\label{sec:soa}

From the viewpoint of Computer Vision, \citet{masood2013computer} provide the only broad review for automated melanoma screening, including 31 works from 1993 to 2012. They propose a general framework for assessing diagnostic models, with quality criteria for the following steps: calibration, preprocessing, segmentation, feature extraction, feature selection, training/testing separation, balancing of positive/negative classes, comparison of results, and cross-validation. In their conclusions, they highlight the challenge of understanding and comparing existing art, promoting an agenda of standard benchmarks and validations, overlooked in previous studies. As we will see, the situation has not --- unfortunately --- improved since.

Our review takes the relay from Masood and Ali Al-Jumaily, with some overlap, focusing on recent papers. We are interested in advanced techniques, in particular, Bag-of-Visual-Words models (\textbf{BoVWs}) and Deep Neural Networks (\textbf{DNNs}) --- two of the main current models for image classification.

\subsection{Image Classification} 
\label{sub:soa-image}

BoVWs and DNNs follow the same information-processing pattern: they extract from the pixels progressively higher-level features, until they can establish the image label. However, they make completely different choices to implement that pattern: BoVWs are  shallow architectures, with three layers of representations (low-level local features, mid-level global features, high-level classifier); DNNs have dozens of layers. BoVWs estimate/learn parameters conservatively: no learning at low level, none to a little unsupervised learning at mid level, and supervised learning only at the high-level classifier. DNNs aggressively learn millions of parameters throughout the entire model.

\citet{sivic03ICCV} and \citet{csurka2004visual} first proposed the BoVW model, appealing to an intuitive notion borrowed from textual information retrieval: the same way a text document can be usefully described as an unordered bag of words, an image could be described as a bag of \emph{visual} words, without concern to geometric structure. That suggestive metaphor helped to popularize the model, although ultimately the parallels between words and visual words were weak. While words are intrinsically tied to semantics, visual words are the result of quantizing feature spaces based upon the \emph{appearance} of images, not their meaning~\citep{AvilaCVIU13}.

Still, the model was very successful. The classical BoVW uses local detectors/descriptors to extract a large amount of discriminating features from small patches all around each image, and then applies an aggressive quantization on the space of those features to establish a visual vocabulary. The vocabulary allows finding visual words that are then aggregated into a frequency histogram. In a less colorful, but more rigorous view, the visual vocabulary is a codebook over the vector space where the low-level features reside, which allows establishing discrete codewords that can be counted. Traditional BoVW ignores all large-scale geometric structure, gaining in generality and robustness but losing in discriminating power. Many extensions appeared, seeking to regain discriminating power without losing generality \citep{LazebnikCVPR06, perronnin2010improving, jegou2010aggregating, picard2011improving, AvilaCVIU13, li2015dictionary}. We will explain the BoVW model and its extensions in more detail in Section~\ref{sec:proposed-bovw}.
    
Deep architectures have a long history, if we accept as deep the model presented by \citet{Fukushima1980}. However, the term \emph{Deep Learning} only caught on after the work of~\citet{Hinton2006}. They published their work at a moment ripe for the aggressive DNN models to thrive, with the availability of large training datasets, and powerful many-core GPU processors. In DNNs, feature extraction and label classification are seamless: deep networks \emph{learn} to extract the features.

The definitive victory of DNNs in Computer Vision came in 2012, when~\citet{Krizhevsky2012} won the classification challenge of the ImageNet Large Scale Visual Recognition Challenge (ILSVRC) 2012. They employed a large, deep convolutional neural network, considerably improving the art at the time.

DNNs have mostly supplanted BoVWs, except in niche applications with little computational power or data. Medical applications, where the training datasets are comparatively tiny, are very challenging for DNNs. Consider that the ImageNet/ILSVRC competition, where DNNs are dominant, has 1.2 million images in the training set; meanwhile, most reported melanoma datasets have less than a thousand images. Nevertheless, with the technique of transfer learning (explained in more detail in Section~\ref{sec:proposed-deep}), DNNs become competitive even in those scenarios.

\subsection{Automated Melanoma Screening} \label{sub:soa-melanoma}

 Literature on automated melanoma screening shows a strong preference for conservative techniques, like the classical BoVW, or even older global color and texture descriptors (Table~\ref{tab:SOTA}). Advanced BoVWs are virtually absent, with the notable exception of a previous work of ours \citep{Fornaciali2014}, which already employed BossaNova with spatial pyramids, and the work of \citet{abedini2015generalized}, which employs soft-assignment coding and spatial pyramids. Very few works attempted DNNs \citep{masood2015self, codella2015deep, premaladha2016novel}. 

\begin{sidewaystable*}
\begin{center}
\begin{footnotesize}
	\caption{A summary of literature on melanoma screening reveals strong preference for conservative Computer Vision techniques. We only included works that deal with lesion classification. Please, refer to Section~\ref{sec:soa} for detailed discussion.}\vspace{-0.2cm}
    \label{tab:SOTA}
    \begin{tabular}{m{3.6cm}m{11.5cm}m{1.2cm}cl}
    \toprule
    \multirow{2}{*}{\textbf{Reference}}                      & \multirow{2}{*}{\textbf{Technique*}}                                                                                                                                & \textbf{Medical†}    & \textbf{~Dataset}            & \multirow{2}{*}{\textbf{Best Result (\%)}}      \\  
                                                             &                                                                                                                                                                     & \textbf{Motivation}  & \textbf{pos~/~neg}                                                             \\ \midrule
	\toprule
	\multirow{1}{*}{\citet{iyatomi08Computerized}}    (2008) & Segmentation. Classical Neural Nets + Global features: Color, symmetry, border, and texture                                                                         & ABCD                 & \multirow{1}{*}{198~/~1060} & auc:92.8 $\cdot$ se:85.9 $\cdot$ sp:86.0         \\ \midrule
	\multirow{1}{*}{\citet{situ08EMBC}}               (2008) & Segmentation. SVM + Classical BoVW. Low level: Wavelets, Gabor-like filters                                                                                         &                      & \multirow{1}{*}{30~/~70}    & auc:82.2                                         \\ \midrule
	\multirow{1}{*}{\citet{diLeo10HICSS}}             (2010) & Segmentation. Decision Tree + Global features: 2D color histogram, Fourier coefficients                                                                             & 7-point              & \multirow{1}{*}{300}‡       & se$>$85.0 $\cdot$ sp$>$85.0     \\ \midrule 
	\multirow{1}{*}{\citet{wadhawan11EMBC}}           (2011) & Segmentation. SVM + Global features: Color histogram, Local Binary Patterns, Haar Wavelet                                                                           & 7-point              & \multirow{1}{*}{110~/~237}‡ & se:87.3 $\cdot$ sp:71.3                          \\ \midrule %
	\multirow{1}{*}{\citet{capdehourat11PRL}}         (2011) & Segmentation. AdaBoost + Global features: 57 shape, color, and texture features                                                                                     & 7-point \& ABCD      & \multirow{1}{*}{111~/~544}  & se:90.0 $\cdot$ sp:85.0                          \\ \midrule 
	\multirow{1}{*}{\citet{wadhawan11ISBI}}           (2011) & Segmentation. SVM + Classical BoVW. Low level: Statistics of 3-level Haar Wavelet decomposition                                                                     & 7-point \& Menzies   & 388~/~912‡                  & auc:91.1 $\cdot$ se:80.8 $\cdot$ sp:85.6         \\ \midrule
	\multirow{1}{*}{\citet{abbas12Skin}}              (2012) & Segmentation. SVM + Global features: Asymmetry, border quantification, color, differential structures                                                               & ABCD                 & \multirow{1}{*}{60~/~60}‡   & auc:88.0 $\cdot$ se:88.2 $\cdot$ sp:91.3         \\ \midrule
	\multirow{1}{*}{\citet{doukas12EMBC}}             (2012) & Segmentation. SVM + Global features: Texture (e.g. ASM and GMSM), size (e.g. area, perimeter), asymmetry index, and color                                           &                      & 800~/~2200                  & acc:85.0-90.0                                    \\ \midrule
	\multirow{1}{*}{\citet{marques12EMBC}}            (2012) & Segmentation. Unspecified Classifier + Global features: RGB, HSV and L*a*b* color histograms; Gradient magnitude and phase histograms for texture                   &                      & \multirow{1}{*}{17~/~146}   & acc:79.1 $\cdot$ se:94.1 $\cdot$ sp:77.4         \\ \midrule %
	\multirow{1}{*}{\citet{mikos12BIOMEP}}            (2012) & Registration. Probabilistic Neural Net + Global features: 12 texture features from pixel-statistics, and 8 from Gray Level Cooccurrence Matrix                      &                      & \multirow{1}{*}{42~/~88}    & acc:69.5                                         \\ \midrule
	\multirow{1}{*}{\citet{barata13iciar}}            (2013) & Segmentation. k-NN + Classical BoVW. Color histograms and moments, Gray Level Cooccurrence Matrix, Gabor filters, Laws Masks and Gradient                           &                      & \multirow{1}{*}{25~/~151}   & se:93.0 $\cdot$ sp:85.0                          \\ \midrule
	\multirow{3}{*}{\citet{abuzaghleh2014automated}  (2014)} & \multirow{3}{*}{Segmentation. SVM + Global features: 2-D FFT and DCT features; size and complexity features}                                                                     &                      & \multirow{3}{*}{40~/~160}   & acc:97.7 (melanoma lesions)                                \\
	& & & & acc:90.6 (normal lesions)                                   \\
	& & & & acc:91.3 (atypical lesions)                             \\
	\midrule   
	\multirow{4}{*}{\citet{barata2014two}            (2014)} & Segmentation. AdaBoost \& k-NN \& SVM. Low level: Gradient histograms; Color histograms over many color spaces;                                                     & \multirow{4}{*}{ABCD}                & \multirow{4}{*}{25~/~151}   & se:96.0 $\cdot$ sp:80.0                          \\ \cmidrule{2-2} 
	                                         & Segmentation. AdaBoost \& k-NN \& SVM + Classical BoVW. Low level: Gradient histograms; Color histograms over many color spaces;                                    &                 &    & se:100 $\cdot$ sp:75.0                           \\ \midrule 
	\multirow{1}{*}{\citet{Fornaciali2014}} (2014) & None. SVM + BossaNova. Low level: RootSIFT                                                                                                                         &                      & \multirow{1}{*}{187~/~560}  & auc:93.7                                         \\ \midrule
	\multirow{1}{*}{\citet{barata2015improving}}      (2015) & Segmentation. k-NN \& SVM + Classical BoVW. Low level: SIFT, 1-D RGB histograms                                                                                     & ABCD                 & \multirow{1}{*}{241~/~241}‡ & se:79.7 $\cdot$ sp:76.0                          \\ \midrule
	\multirow{2}{*}{\citet{abedini2015generalized}   (2015)} & Segmentation. SVM + BoVW soft assignment. Low level: Color, edge, color LBP histograms, SIFT                                                          &                      & \multirow{1}{*}{40~/~160}   & ap:92.7 $\cdot$ se:90.0 $\cdot$ sp:90.0          \\ 
	                                        &   and color SIFT                                                                                                                                                                  &                      & \multirow{1}{*}{391~/~2761} & ap:96.7 $\cdot$ se:98.7 $\cdot$ sp:94.8          \\ \midrule
	\multirow{1}{*}{\citet{ohki2015building}}         (2015) & Segmentation. Random Forest + Global features: Top 24 features chosen from 120 color features                                                                       & ABCD                 & \multirow{1}{*}{168~/~980}  & se:79.8 $\cdot$ sp:80.7                          \\ \midrule
    \multirow{1}{*}{\citet{masood2015self}}           (2015) & Segmentation. Deep Neural Net \& SVM + Global features: Texture features --- autoregressive model, Gray Level Cooccurrence Matrix, Fuzzy-Mutual Information Wavelet &                      & \multirow{1}{*}{120~/~170}  & acc:89.0 $\cdot$ se:90.0 $\cdot$ sp:88.3         \\ \midrule
	\multirow{1}{*}{\citet{codella2015deep}}          (2015) & SVM + BoVW with Sparse Coding and features extracted from Deep Neural Net pretrained on ILSVRC 2012                                                                 &                      & \multirow{1}{*}{334~/~2290} & acc:93.1 $\cdot$ se:94.9 $\cdot$ sp:92.8         \\ \midrule
	Premaladha and\textcolor{white}{Ravichandrana} Ravichandran \citep{premaladha2016novel}      (2016) & Segmentation. Deep Neural Net \& Hybrid AdaBoost + Statistical and shape features (e.g. area, perimeter, asymmetry index)                                           &                      & \multirow{1}{*}{992}        & acc:92.9                                         \\ 
    \bottomrule
    \end{tabular}
\vspace{0.1cm}\\
    \begin{tablenotes}
    \centering
    \item * Described as (omitting absent steps): Preprocessing. Classifiers + (Mid-Level/Global) Features. Local features.
    \item † Only when authors explicitly motivate their method on medical criteria. | ‡ Subset from EDRAS Interactive Atlas (Section~\ref{sec:dataset}).
    \item acc: accuracy | ap: average precision | auc: area under the ROC curve | se: sensitivity | sp: specificity  
    \end{tablenotes}
\end{footnotesize}
\end{center}
\end{sidewaystable*}


Current lesion classification tends to follow a three-step scheme: (i) segment the lesion, (ii) extract features from the lesion-region only, and (iii) classify the lesion. Often, a preprocessing step appears before segmentation to reduce noise, and, sometimes to eliminate artifacts. 

Almost all methods impose segmentation. In our review, the only exceptions were the works of \citet{mikos12BIOMEP} and \citet{codella2015deep}. The prominence of segmentation is such, that some works deal exclusively with it \cite{celebi2009lesion, silveira2009comparison, sadri2013segmentation}. We cite those works for the sake of completeness, but do not detail them further, since we limit our scope to lesion classification. We refer the reader interested in lesion segmentation to the work of \citet{silveira2009comparison} and of \citet{celebi2009lesion}, who, together, survey a large number of techniques.

Segmentation also appears prominently in works whose main objective is classification. \citet{wadhawan11ISBI} evaluate three segmentation algorithms: ISODATA, fuzzy c-means and active contour without edges, finding the latter performs best. \citet{abuzaghleh2014automated} propose a novel segmentation based upon Gaussian filters. \citet{capdehourat11PRL} find no clear winner among several segmentation choices, and thus prefer the simple and fast color-based Otsu method.

Due to the accent on segmentation, the presence of artifacts, either natural (hair, veins), or artificial (rulers, masks, bandages, etc.) is a major obstacle for most techniques. \citet{abbas12Skin} deal with hair removal explicitly, using the Derivative of Gaussian filter, morphological operations, and fast matching. \citet{wighton2011generalizing} proposed an integrated framework that can be used for many preprocessing tasks: segmentation, hair detection and identification of dermoscopic structures (pigment network). They base their method on supervised learning, using a multivariate Gaussian model that labels each pixel, using maximum a posteriori estimation.

As mentioned, literature favors conservative global color and texture descriptors for feature extraction. 
Global descriptors, favored by most approaches skip the mid level altogether, and move directly to classification. 
Color histograms and color moments are the most employed color descriptors \citep{wadhawan11EMBC,situ08EMBC,barata13iciar}. Texture descriptors vary more: Haar wavelets \citep{wadhawan11EMBC,wadhawan11ISBI}, Gabor filters \citep{situ08EMBC,barata13iciar}, Gray Level Cooccurrence Matrix \citep{mikos12BIOMEP,doukas12EMBC}, etc. Most studies \citep{abedini2015generalized,barata2014two,marques12EMBC,iyatomi08Computerized} reported best results with combined color and texture descriptors.

When mid level appears, it is virtually always the classical BoVW approach, with hard assignment, sum/average pooling, and small codebooks of up to 500 codewords (we will explain those terms in Section~\ref{sec:proposed}). \citet{wadhawan11ISBI,situ08EMBC,barata13iciar,barata2015improving} follow that classical BoVW path, using traditional color and texture features as low-level local features. We will explore in depth the work of \citep{wadhawan11ISBI}, chosen as baseline for our comparisons, in Section~\ref{sec:baseline}.

Most works favor Support Vector Machines (\textbf{SVM}) \citep{vapnik95}, a robust technique with easy-to-use, readily available implementations. A couple of other classifiers appear, such as neural networks --- in their classical shallow incarnation --- \citep{iyatomi08Computerized,mikos12BIOMEP}, decision trees \citep{diLeo10HICSS,capdehourat11PRL}, or random decision forests~\citep{ohki2015building}.

Recently, DNNs and other Deep Learning architectures started to appear as novel methods for automated melanoma screening \cite{masood2015self, codella2015deep, premaladha2016novel}. We give special attention to those works.

\citet{masood2015self} present a semi-supervised model --- i.e., built from both labelled and unlabelled data --- based upon a deep belief network combined with SVM. They preprocess the images removing hair and artifacts,  and segment the lesion with a histogram-based fuzzy c-means thresholding algorithm. Texture features are extracted with autoregressive models, Gray Level Cooccurrence Matrix, and Fuzzy-Mutual Information based Wavelet packet Transform. Those features feed three independent classifiers: a deep belief net, a self-advised SVM with RBF kernel, and another SVM with polynomial kernel. Classifiers outputs are weighted by least-square estimation into a final decision. Classification error decreases as the amount of labeled data grows; still, in their comparison, the proposed model makes the most efficient use of unlabeled data.

\citet{codella2015deep} combine deep learning and sparse coding for feature extraction, with SVM for classification. They employ both a DNN pretrained on ImageNet, and an enhanced BoVW based on sparse coding and average pooling over both grayscale and color images. They employ non-linear SVM, with histogram-intersection kernel and sigmoid feature normalization. They also evaluate fusion of the classifiers opinions (late fusion), finding some improvement when all attempted models are fused. This work is the most similar to our own proposal at DNN with transfer learning (Section~\ref{sec:proposed-deep}).

\citet{premaladha2016novel} propose a complex framework for automated melanoma screening, preprocessing images with Contrast Limited Adaptive Histogram Equalization and median filter to filter out noise (like hair and scars); segmenting lesions with the normalized Otsu's algorithm; extracting fifteen different features based on pixel statistics (mean, standard deviation, entropy, correlation, etc.) and lesion shape (area, perimeter, diameter, asymmetry index, fractal dimension, etc.); and finally classifying with Deep Learning, and Hybrid AdaBoost SVM. 

Virtually always in the papers we surveyed, the experimental protocol is some form of cross-validation. With few exceptions, due to the small sizes of the datasets, experimenters favor many folds, from 10 folds until leave-one-out schemes \citep{iyatomi08Computerized,mikos12BIOMEP,marques12EMBC}.

    \subsection{Criticism of Existing Art} 
\label{sec:problems} 
Existing art in automated melanoma screening have important shortcomings. The most severe are outdated Computer Vision techniques, and inadequate --- and irreproducible --- experimental validation. The latter issue is not exclusive to melanoma screening, but happens on all computing literature \citep{peng2011science,sandve2013plos}, and even scientific literature in general \citep{allison2016reproducibility}.

Existing art often attempts to directly mimic medical diagnostic rules, especially
the ABCD Rule of Dermoscopy~\citep{ABCDrule}, and the
7-Points Checklist~\citep{argenziano98Arch}.

Attempting to mimic human procedures leads to rigid processing pipelines based on segmentation $\rightarrow$ feature extraction $\rightarrow$ classification, which no longer reflect the state-of-the-art in Computer Vision. That is compounded by the use of features imitating medical reasoning (color, texture),  which are known to underperform when compared to recent options.

The choice to segment is among the most questionable. Intuitively, it makes sense: segmentation is needed to analyze lesion shape, or to isolate the lesion from healthy skin and focus only on the former. In addition, early models of Computer Vision imposed segmentation before complex image understanding, due to the classical vision model of David Marr~\citep[chapter 4]{marr1982vision}. Such view, however, was abandoned with the remarkable success of local features, which allowed very robust image matching, tracking, and depth reconstruction without any segmentation~\citep{li2008comprehensive,tuytelaars2008local}. The success of local features was a prelude to the BoVW model, in which local features are employed in aggregate form, also without preliminary segmentation. The DNNs that supplanted the BoVW model forgo pre-segmentation as well; when segmentation appears in current DNNs, it is as a goal in itself, considered harder than classification~\citep{Everingham10,ILSVRC15}.

Even if literature were successful in imitating the ABCD Rule, or the 7-Points Checklist, one could still argue that those rules were supplanted by the newer 3-Point Checklist~\citep{soyer2004three}, and the Revisited 7-Point Checklist~\citep{argenziano2011seven}. More telling is the fact that none of those rules ``caught on'' among doctors: traditional pattern analysis~\citep{pehamberger1987vivo} is still considered the most reliable technique to teach dermoscopy~\citep{carli2003pattern}.

It is refreshing to see novel works in melanoma screening employing DNNs. Still, it is interesting to note that, of the three works, two \citep{premaladha2016novel, masood2015self} employ DNNs still in the context of complex, traditional pipelines, based upon segmentation, preprocessing, etc. The work of \citet{codella2015deep} marks, in that sense, a more definitive depart towards modern Computer Vision: a streamlined pipeline, with a modern BoVW on one hand, and a DNN+transfer learning on the other hand. Our own proposals will follow the same trajectory.

All those issues, however, pale in comparison to the problem of validation and reproducibility. Despite the existence of more than a dozen published papers on the subject, there is little reliable accumulated knowledge.

Authors choose different metrics: most only report global accuracy --- which is misleading when the positive and negative classes are highly unbalanced. If a dataset has 20 images of melanoma and 80 images of non-melanoma --- a disproportion quite typical of melanoma datasets --- one can classify all images as non-melanoma and still get 80\% accuracy. Therefore, 
some authors prefer the pair Sensitivity--Specificity, which together, reveals such problems (in the example above, one would get 100\% specificity, but 0\% sensitivity). Some authors --- us included --- prefer to report the area under the ROC curve (\textbf{AUC}), which averages all possible Sensitivity--Specificity pairs given by the classifier, and measures, so to speak, the amount of ``knowledge'' the classifier has about the problem, without regard to its preference for sensitivity or for specificity. More exotic metrics, like the F-measure of information retrieval, also appear. 

But agreeing on metrics is a necessary, not sufficient, condition to render the works comparable.

Many works are validated in datasets which are simply too small to provide reliable evidence. Half the papers we surveyed employ datasets of 300 images or less. Further, those datasets are nearly always proprietary, and not available for cross-examination. (We further discuss the issues of selecting datasets and choosing class labels in Section~\ref{sec:dataset}.) Comparison would still be possible, in principle, if each research group ran previous techniques in their own data, but authors will seldom share code.

Attempts at comparison, thus, must incur in the expensive effort of \emph{reimplementing} previous works. But even that is often impossible. The vast majority of works is simply not described in a level of detail that allows reimplementation. That fact was already observed by \citet{masood2013computer}, in works that spanned from 1993 to 2012; our exam of works from 2008 to 2015 found a situation just as dire. We could not contact all works listed in Table~\ref{tab:SOTA}, so --- protecting innocent and guilty alike --- the reader should not judge any particular author. But the fact remains that we contacted a large number of authors and: mostly got no answer at all; or --- rarely --- got an explanation that they could not share neither data nor code, and sometimes, not even details about their methods. We feel morally obliged to acknowledge the exceptions: \citet{silveira2009comparison} shared their data, and \citet{wighton2011generalizing} shared their code with us without imposing any obstacles (those works do not aim directly at diagnosis, could not be compared to ours, and do not appear in Table~\ref{tab:SOTA}). \citet{abbas12Skin}, generously shared their annotations on the dataset we employed.

Due to the current complexity of the computer systems employed in scientific experiments, the reproducible paper without original code is a unicorn, a mythical creature. Even the most detailed description leave out important information, to respect page limits, or to avoid boring/confusing the reader with a  myriad of minutiae. Still, a single wrong parameter can collapse an entire experimental cathedral.

We started our efforts to reproduce existing art on mid 2014. At the time, the only work we could find, whose description gave us any hope to reproduce was by~\citet{wadhawan11ISBI}. The attempts to reimplement their method are described in Section~\ref{sec:baseline}. The problems we faced were typical: missing details, results much below the ones reported originally, and --- despite insistent attempts --- failure to earn the cooperation of the author. Still, although the light we shed at their work might be unflattering, the reader should consider that \emph{theirs was the only paper we found that allowed even that minimal attempt at reproduction}. So, in terms of reproducibility --- despite the mismatch between the numbers they report and the ones we found --- their work is still the \emph{best} we found in the literature of automated melanoma diagnosis! \\

\noindent\textbf{Joint efforts and new developments.} One can find tentative skin-lesion datasets in the Web, but often those datasets lack size, quality, or reliable annotations. We are aware of only two efforts worth mentioning: the PH² Dataset\footnote{www.fc.up.pt/addi/ph2\%20database.html}~\citep{datasetMendonca}, and the ISIC Project\footnote{www.isdis.net/isic-project}. 

The PH² Dataset has 200 dermoscopic images (80 common nevi, 80 atypical nevi, and 40 melanomas), acquired at the Dermatology Service of Hospital Pedro Hispano/Portugal, providing also ground truths for segmenting the lesions.  \citet{marques12EMBC,barata13iciar,abuzaghleh2014automated,barata2014two} all report experiments on it.

The International Skin Imaging Collaboration: Melanoma Project (ISIC) is an academia/industry partnership, coordinated by the International Society for Digital Imaging of the Skin, to acquire and annotate skin lesion images. As of March 2016, $\sim3000$ images were available, moving through a multi-phase quality-control pipeline, but no image had passed ``Phase 2'' (concerned with annotation quality according to summary descriptions available at their website). Nevertheless, most images provide at least diagnostic labels, with many providing much more. Although we believe ISIC will soon become an important standard dataset for the community, we are aware of only two published works reporting experiments on it: \citet{abedini2015generalized,codella2015deep}.

ISIC promoted the first open challenge for automated skin lesion annotation, in three phases: lesion segmentation, dermoscopic feature extraction, dermoscopic feature segmentation, lesion classification, and segmented lesion classification. They provided $\sim900$ training images, with ground truths, and asked for results on $\sim350$. We became aware of the challenge too late to effectively participate: we could barely download and study the data. Nevertheless we are excited by that new development in the community, and advocate public challenges --- with testing labels unknown to the participants --- as one way to improve experimental reliability when researchers are forced to work with small datasets (see the final discussion).




    \section{Proposed Solutions} 
\label{sec:proposed}
We introduce two novel approaches for automated melanoma screening, based in modern Computer Vision techniques: advanced BoVWs, and DNNs with transfer learning. A reference implementation for our techniques (and for our reimplementation of the baseline technique, described in the next section) is also available\footnote{https://sites.google.com/site/robustmelanomascreening/}. The reference code also contains the scripts and explanations needed to redo the experiments.

\subsection{BossaNova for Melanoma Screening }
\label{sec:proposed-bovw}

\begin{figure*}[th]
    \centering
	\includegraphics[width=\linewidth]{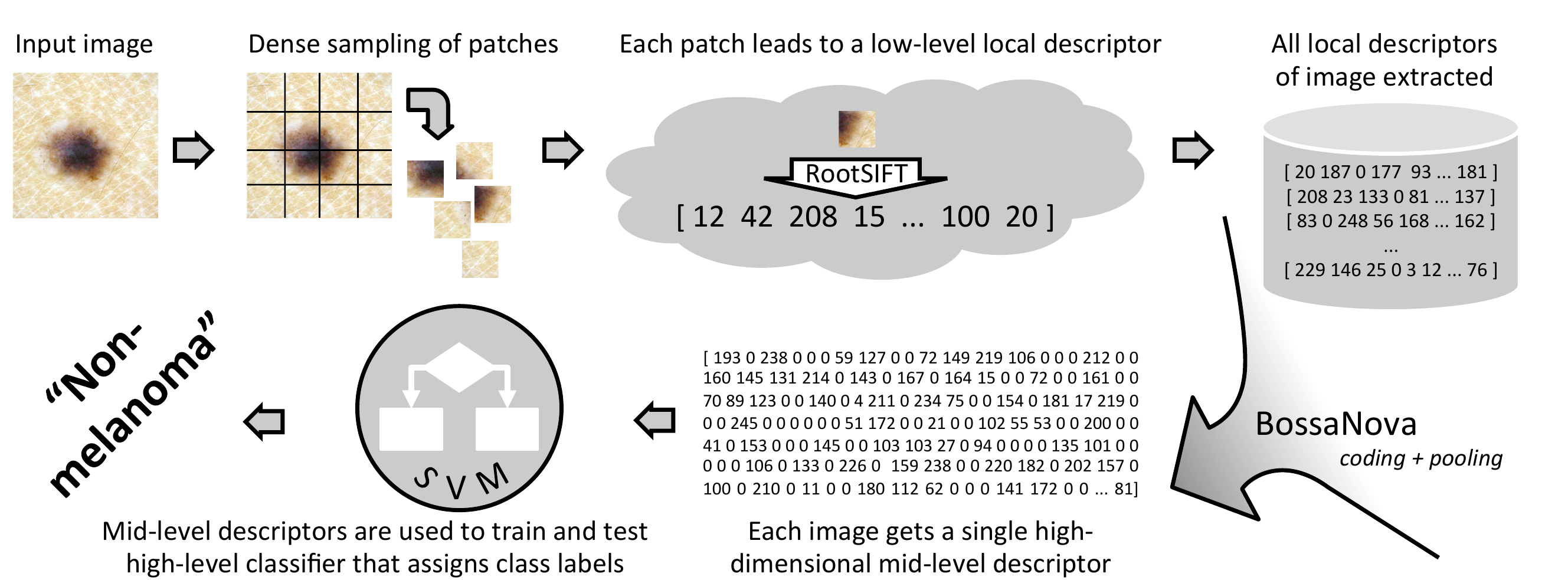} 
		\caption{The main pipeline of BossaNova, an enhanced Bags-of-Visual-Words (BoVW) model. Like all BoVWs, BossaNova moves from the input image towards the class label in a three-tiered process: (1) low-level feature extraction, which extract robust local features from patches densely sampled around the image; (2) mid-level feature extraction from the local features; (3) high-level classifier. BossaNova's main novelty is the mid-level coding/pooling scheme.}
	\label{fig:pipeline}
\end{figure*}

As explained in Section~\ref{sub:soa-image}, the BoVW model was prevalent before the introduction of DNNs. It still finds niche applications where data or computational resources are too scarse for the optimal application of DNNs. 

Here, we abandon the ``visual word'' metaphor, and adopt the more formal and recent point of view that the BoVW is an advanced \emph{encoding} of local features~\citep{boureau10CVPR,AvilaCVIU13}. Conceptually, the BoVW mid-level features are obtained by extracting low-level local features and submitting them to a \emph{coding} and a \emph{pooling} functions.

The theory of discriminating low-level local features was mainly developed during the early 2000's~\citep{schmid2000ijcv,mikolajczyk2005performance}, reaching maturity with the SIFT feature detector and extractor~\citep{LoweIJCV04}, which became the gold standard. Local features extract robust information over small patches of the images. They have much discriminating power, and allow matching target object and scenes with great accuracy --- but are not able to \emph{generalize}, and identify \emph{classes} of objects. We now understand the BoVW model as a mechanism to provide that generalization power.

The most important operation to obtain generalization is the \emph{pooling}, which flattens all local features into a single feature vector of fixed  --- but frequently huge --- dimensionality. Conceptually, the pooling marginalizes over many of the transformations for which we want invariance: translations, rotations, etc. However, a naïve pooling --- e.g., the simple averaging of local features --- is extremely destructive.

The \emph{coding} operation prepares the local features to be pooled and still preserve relevant information. In the classical BoVW, the pooling is an average operation, which, for example, over SIFT features without coding, would just obtain the uninformative average SIFT. Classical coding uses a large codebook by taking the centroids of a $k$-means over a sample of training local features. It then maps each local feature into a $k$-dimensional vector, using one-hot encoding, where the hot dimension corresponds to the closest centroid to the input vector. That coding is called \emph{hard assignment}, because of the sharp distinction between the dimensions with a 0 and those with an 1 in one-hot encoding. Hard assignment is, itself, destructive, but when paired with average-pooling  leads to a richer mid-level feature vector: a histogram of the frequency of each ``group'' of local features, i.e., a histogram of the \emph{local appearances} of the image.

That classical BoVW --- hard assignment + average pooling --- stills throws away a lot of information. In particular, all geometric information is lost. \citet{LazebnikCVPR06} proposed the enhanced multi-level pooling, \emph{Spatial Pyramid Matching}, to recover some large-scale geometric information. Other extensions offered novel coding and pooling operations to preserve more information from the local features. 

\citet{AvilaCVIU13} proposed BossaNova as an enhanced mid-level, with a density-based pooling strategy. BossaNova improves BoVW, employing, instead of hard assignment, a \emph{localized soft assignment} that associates the local features to the codewords according to the distance between them. BossaNova also introduces a \emph{density-based pooling}: instead of simply counting frequencies like  ordinary average-pooling, BossaNova traces a compact local distribution of the distances between local features and codewords --- a mini-histogram of distances.


A complete redescription of BossaNova is beyond the scope of this paper. We refer the reader to \citet{AvilaCVIU13}, and to the reference implementation\footnote{https://sites.google.com/site/bossanovasite}. 

BossaNova was successfully employed for computer-aided diagnosis, for diabetic retinopathy \citep{pires2014automatic}, and for melanoma itself, in a previous work of ours \citep{Fornaciali2014}.

In our proposal, the BossaNova pipeline works as follows:
\begin{enumerate}
\item All images are resized to 100K pixels, if larger, using ImageMagick's convert command (version~6.8.6-10).
\item The low-level local features are extracted from all images and stored. We employ dense RootSIFT~\citep{RootSIFT}, implemented in VLFeat toolbox~\cite{vedaldi10vlfeat} (version~0.9.17), with a grid step of $8$ pixels and patch sizes of $12, 26, 58$ and $128$ pixels.
\item To sparsify the low-level features, all dimensions below $0.0025$ are mapped to zero.
\item A set of 500K distinct descriptors is randomly sampled from the training set to facilitate the PCA and codebook learning that follows.
\item 
We learn a PCA matrix in the 500K subsample above. We learn and store the mean of the subsample, center the features using that mean and apply a PCA procedure implemented in Java. 
The PCA decorrelates, and reduces the dimensionality from $128$ to $64$. Our PCA is non-whitening, i.e., it does not erase the variances along the axes. Using the computed mean and matrix, we propagate the PCA to all other local features.
\item We learn the codebook by picking at random $k=2048$  distinct features from the PCA-transformed 500K subsample.
To easily select $k$ features at random, we perform a single iteration of a $k$-means routine implemented in Java. 
\item BossaNova creates the mid-level features from the PCA-transformed low-level features. The reference implementation is employed, with number of bins $B=4$, and minimal and maximal range of distances $\lambda_{min}= 0.6$ and $\lambda_{max}= 1.6$. Lazebnik's Spatial Pyramid Matching~\citep{LazebnikCVPR06} is employed at the end, with 5 regions ($1\times1$, $2\times2$) --- they are built-in in BossaNova's reference code.
\item The output of BossaNova forms the mid-level features employed in the training and testing of the high-level decision model. We employ an SVM classifier with an RBF kernel, implemented in LIBSVM~\citep{chang2011libsvm} (version~3.17). 
\item SVM margin hardness $C \in \{2^c: c\in[-5, -3, \ldots, 15]\}$ and kernel width $\Gamma\in \{2^\gamma: \gamma\in[-15, -13, \ldots, 3]\}$ are learned by internal cross-validation on the training set, using the ``easy.py'' script of LIBSVM, which selects the best value. We slightly modified easy.py to select the model with best AUC instead of the model with best accuracy. 
\end{enumerate}

The experimental procedure was common to all tested methods, and will be explained at Section~\ref{sec:experimental}. The whole scheme is illustrated in Figure~\ref{fig:pipeline}.

Many of the seemingly arbitrary choices (sparsification of the low-level features, decorrelation with PCA, use of random codebook instead of using expensive clustering) reflect the accumulated wisdom learned by the Computer Vision community during the advancement of the BoVW model. Most of that wisdom is nicely documented in the papers of~\citet{chatfield2011devil} and \citet{akata2014good}. In addition, a preliminary work of ours \citep{Fornaciali2014} focused on fine-tuning BossaNova for melanoma screening: we kept in the pipeline above most of the best choices found in that previous work. 

\subsection{Deep Architectures with Transfer Learning}
\label{sec:proposed-deep}

Deep Learning for visual tasks usually involves seamlessly learning every step of the classification process, from feature extraction to label assignment. That pervasive learning improves generalization, but brings its own challenges: DNNs estimate a huge number of parameters, requiring large amounts of annotated data and computational resources.

DNNs are based on Artificial Neural Networks, which have a long history, the first ideas dating back to 1943, when McCulloch and Pitts~\cite{mcculloch1943logical} proposed a mathematical model for the biological neuron. However --- as we did with the ``words'' in ``visual words'', we caution the reader against the ``neural'' in ``neural networks''. The similarity of modern DNNs to biologically plausible neural models is questionable, to say the least. Today, the justification for DNNs rests not on biological metaphors, but on their ---  empirically validated --- amazing performance in a broad variety of tasks. 

DNNs may also be approached as multi-layered generalized linear models learned by maximum likelihood, without need of biological justifications, and that is the viewpoint we adopt here. On the simplest DNNs, each neuron $i$ on layer $\ell$ may be described by the function $f_{i,\ell}(x):\mathbb{R}^{n_{\ell-1}}\rightarrow\mathbb{R}=\psi_\ell(W_{i,\ell}\cdot x_{\ell-1}+b_{i,\ell})$, where the input vector $x_{\ell-1}$ gathers the output of all neurons on the previous layer $\ell-1$ (the input image, if $\ell=1$), the vector $W_{i,\ell}$ and scalar bias $b_{i,\ell}$ are the weights of the neuron --- the values that are effectively learned during training, and $\psi_\ell$ is a fixed (not learned) non-linear function, often a sigmoid or a ramp function. On the DNNs employed for image classification the description is a little more complex because the lower neurons are based on \emph{convolutions}, but the fundamentals do not change. We refer the reader to \citet{LeCun2015} for a complete description.

The weights are learned by minimizing the error of the network predictions over the training set, a highly non-convex optimization that must be approached numerically, often with gradient descent methods. The gradient computations are made feasible by a technique called \emph{backpropagation}, a clever way to use the chain rule for computing the huge derivative of the prediction error. The whole procedure, as it finds the weights that best adjust the model to the training set, is conceptually similar to a maximum likelihood estimation. As such, large models require huge training sets, and still are liable to under or overfitting. The craftsmanship of DNNs is delicate, full of ad-hoc ``tricks of the trade'' for regularizing and optimizing the models.

The greediness of DNNs for labelled data makes them cumbersome for medical applications, due to the costs of acquiring such data. As we mentioned, DNNs thrive with hundreds of thousands, up to several millions learning samples, while a medical dataset of a thousand samples is considered large!

That difficult is, however, addressed by \emph{transfer learning}, a technique that allows recycling knowledge from one task to others. In visual tasks, the low-level layers of a DNN tend to be fairly general; specialization increases as we move up in the networks \cite{Yosinski2014}. That suggests we may learn the general layers on \emph{any} large dataset, and then learn only the specialized upper layers on the small medical dataset.

\begin{figure}[tb]
    \centering
	\includegraphics[width=\linewidth]{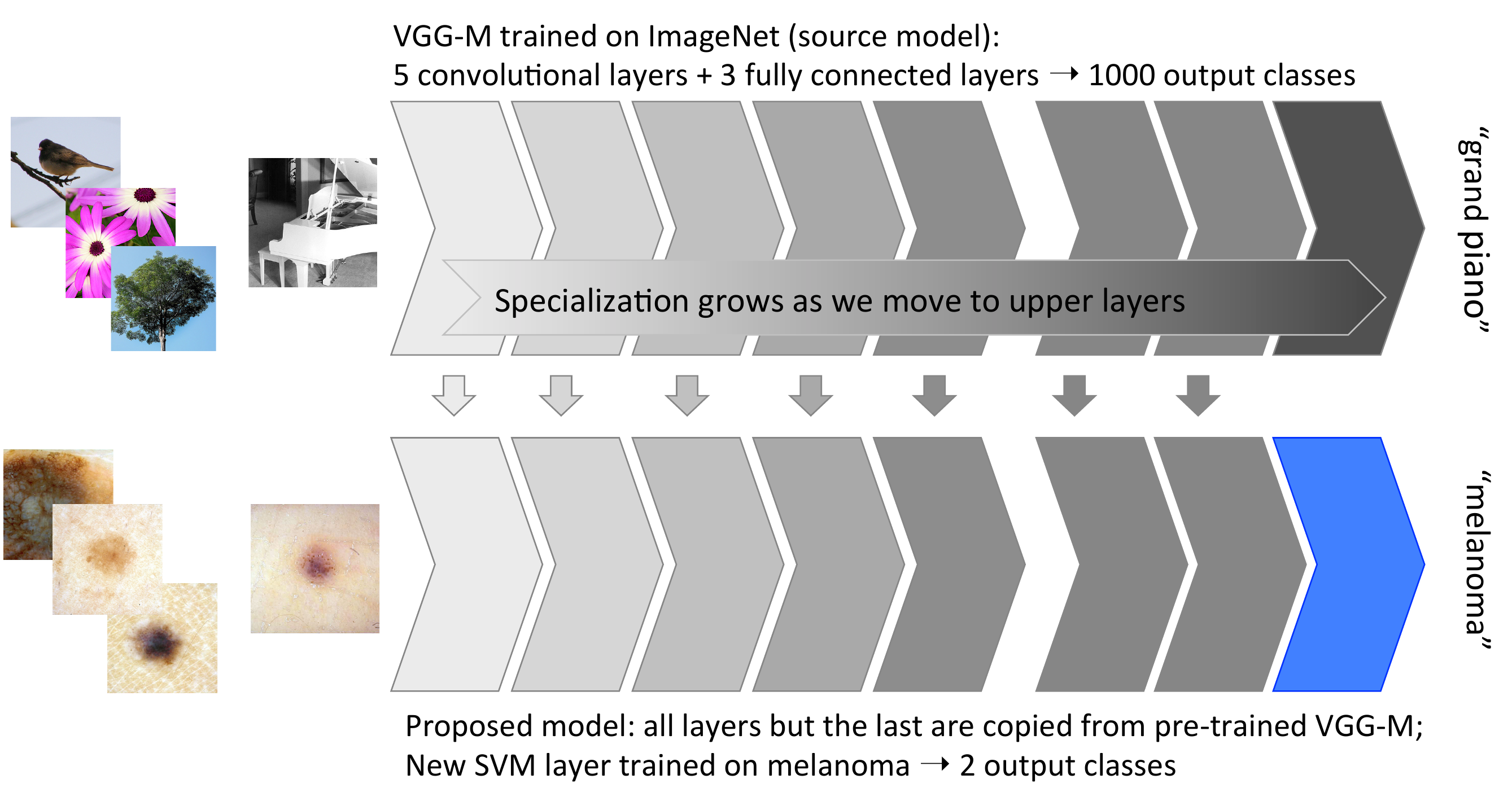}
	\caption{The proposed melanoma screening classifier using Deep Learning and Transfer Learning. We transfer knowledge from a Deep Neural Network pretrained on ImageNet (top row).  All but the topmost layers are used to extract features that are fed to a new SVM layer, which is trained for the new task of melanoma screening.}
	\label{fig:deep-transfer}
\end{figure}

A straightforward strategy for transfer learning is freezing the weights of a pretrained deep neural network up to a chosen layer, replacing and retraining the other layers for the new task (Figure~\ref{fig:deep-transfer}). The new ``layers'' do not have to be themselves neural: other classifiers are perfectly acceptable. 

More sophisticated techniques, which fine-tune the lower layers instead of freezing them, are available \citep{Yosinski2014}. That means allowing those layers to evolve, expecting them to slightly adapt to the new task. That tends to correct any biases learned from the original task. When fine-tuning, it is easier to keep the entire network neural, to simplify the computation of the error gradients.

Vanilla transfer learning, without fine-tuning, is the simplest and cheapest --- and allow easily plugging an SVM classifier on the top layer. Thus, it is the one we evaluate in this work. In our pipeline, we transfer knowledge from a  pretrained MatConvNet~\citep{matconvnet} model proposed by Chatfield~et~al.~\citep{Chatfield2014}, freezing all layers inside the blue-dashed rectangle in Figure~\ref{fig:deep-transfer}. The output of layer 19 is fed to an SVM classifier, which we train for the new task.

The procedure is illustrated in Figure~\ref{fig:deep-transfer} and detailed below:

\begin{enumerate}
\item We obtain the pretrained MatConvNet \textit{VGG-M} model\footnote{http://www.vlfeat.org/matconvnet/pretrained/}. The pretrained network is a .mat file with a complete description of the network, and all layer weights pretrained for the ImageNet task. 
\item All images are resized to exactly $224\times 224$ pixels, as expected by the model, using a bicubic interpolation. The images whose aspect ratio is not square are distorted to fit.
\item The model-input mean image is subtracted from all input images, to ``centralize'' them. The input mean is computed during the training of \textit{VGG-M} in ImageNet and is stored in the .mat file as part of the model. 
\item The centralized images are fed to the pretrained \textit{VGG-M} model.
\item The outputs of Group 7/Layer 19, vectors of $4096$ dimensions, are obtained, $\ell_2$-normalized and used as mid-level features.
\item The mid-level features are employed in the training and testing of the high-level decision model, an SVM classifier with linear kernel, implemented with LIBLINEAR~\citep{fan2008liblinear} (version~1.96).
\item SVM margin hardness $C\in \{10^c:c\in[-4:3]\}$ is learned by internal cross-validation on the training set, by selecting the value that maximizes the AUC.
\end{enumerate}

    \section{Baseline}\label{sec:baseline}
As explained in Section~\ref{sec:problems}, finding a suitable baseline was extremely difficult, due to authors withholding data and code. When we acquiesced to reimplement some technique from scratch, the work of \citet{wadhawan11ISBI} was the only one, at the time, at the intersection of these criteria:
\begin{itemize}
  \item Good performance: the authors report an AUC of 91.1\%;
  \item Validation on a sufficiently large dataset: 1\thinspace300 images (388 positive). Better yet, they employ a well-known, commercially available dataset, which we were able to secure (Section~\ref{sec:dataset});
  \item Description sufficiently detailed and self-contained to allow \emph{considering} reimplementation.
\end{itemize}

Strictly speaking, the technique of Wadhawan et al. is a form of Classical BoVW, using hard assignment and sum pooling. Sum pooling is closely related to average pooling, with the resulting vector counting \textit{occurrences} instead of \textit{frequencies}. Instead of invariant low-level features, like SIFT or RootSIFT, they employ custom, low-dimensional local features, based upon statistics of a three-level Haar wavelet decomposition of the image.

Wadhawan et al. test three segmentation algorithms, reporting best results with Active Contours without Edges~\citep{chanvese2001} --- the one we chose to reproduce. They omit, however, all implementation details about the segmentation. We employed code provided by Yue Wu\footnote{Available at MATLAB's repository: \url{mathworks.com/matlabcentral/fileexchange/23445-chan-vese-active-contours-without-edges/content/chenvese.m}}. Wu provides three variants, from which we chose the default `chan'. We pick by trial the cost of the perimeter $\mu=0.25$. We boot up the segmentation mask with a disk in the center of the image, with a radius of $0.25\times p$, where $p$ is the smallest of the height or width of the image, and let it evolve for 2000 iterations.

The detailed description follows. In the text below, Roman text indicates information given by \citet{wadhawan11ISBI}, or by some paper they cite. \textit{Italic} text indicates information we had to complete to the best of our knowledge. 

\noindent\textbf{Preprocessing.} The \textit{full size} images are converted to 256-level grayscale.

\noindent\textbf{Segmentation.} Each input image is segmented to extract the lesion region:
    \begin{enumerate}
    \item The input for segmentation are the images filtered by a median kernel, \textit{of size corresponding to $5\times p/256$ ($p$ like above). (The original images, without filtering, are employed for feature extraction)}.
    \item Active Contours (detailed above) outputs the segmentation binary mask.
    \item Morphological operators of opening and closing are applied to remove small regions on the mask. \textit{We employed one opening, to get rid of small components, followed by one closing, to get rid of small gaps. The structuring element was always a flat disk of radius $3\times p/256$}.
    \item If more than one component remains, we keep the one with larger area.
    \end{enumerate}
\noindent\textbf{Low-level} local feature extraction:
    \begin{enumerate}
    \item We select square patches centered at every $10$ pixels, with a side of $24$ pixels, but only those inside the lesion mask \textit{(those whose center falls inside the lesion mask).}
    \item For each patch we compute its 3-level Haar wavelet decomposition.
    \item For each patch we compute an independent local feature formed by the mean and standard deviation of each sub-band of the decomposition. \textit{There are 3 sub-bands for each level (vertical, horizontal, and diagonal), plus the final ``approximation'' sub-band. The feature vector has, thus, 20 dimensions.}
    \end{enumerate}
\noindent\textbf{Mid-level} feature extraction:
    \begin{enumerate}
    \item We learn the mean and the standard deviation of each dimension on the training set, and use them to z-normalize all local features.
    \item We apply a $k$-means procedure \textit{on a maximum subsample of 1M} of the feature vectors, with $k=200$, and employ the $k$ centroids as codebook. \textit{We leave $k$-means turn for a  maximum of $1000$ iterations.}
    \item For each image, we initialize a mid-level feature vector of size 200 as 0. We  take each of the image's local features, check the closest centroid, and increment by~1 the corresponding dimension on the mid-level features.
    \end{enumerate}
\noindent\textbf{High-level model} learning: the mid-level features are used as input to train and test an SVM classifier. \textit{For fairness, we employ the same SVM procedure explained in Section~\ref{sec:proposed-bovw}.} 

Insistent attempts were made to contact the faculty author --- the students' contacts were no longer valid ---  to obtain the missing details, even more so after we were unable to reach the numbers they report. We only got two short answers, one at the beginning of our attempts claiming secrecy due to intellectual property restrictions, and one much later in the processes alleging lack of time. In the end, we were not even sure our datasets matched, since we knew, due to the number of images they report, that they had made some unspecified  selection over the data. A final attempt was made prior to the submission of this paper to review, with a full copy of our reimplemented code, but it also went unanswered.

The spirit here is not to single out one author for condemnation --- as far as we know, had we chosen another work, we could have faced the same problems --- but to highlight how current practices on the topic of melanoma screening invert scientific burden-of-proof, charging a disproportionate cost on the authors who are trying to replicate results. We develop this topic further at the discussion in the end of the paper.

\section{Experiments} \label{sec:experimental}

\subsection{Dataset}
\label{sec:dataset}

Finding a suitable dataset is the main challenge for research on machine learning for medical applications. Large, high-quality, publicly accessible (even under payment), reliably annotated datasets are as rare as they are valuable.

The Interactive Atlas of Dermoscopy~\citep{argenziano2002dermoscopy} appears as an attractive possibility: a multimedia guide (Booklet + CD-ROM) intended for training medical personnel to diagnose skin lesions. It has 1000+ clinical cases, each with at least two images of the lesion: close-up clinical image (acquired with a Nikon F3 camera mounted on a Wild M650 stereomicroscope), and dermoscopic image (acquired with a Dermaphot/Optotechnik dermoscope). Most images are 768 pixels wide $\times$ 512 high. Each case has clinical data, histopathological results, diagnosis, and level of difficulty. The latter measures how difficult (low, medium and high) the case is considered to diagnose by a trained human. The diagnoses include, besides melanoma (several subtypes), basal cell carcinoma, blue nevus, Clark's nevus, combined nevus, congenital nevus, dermal nevus, dermatofibroma, lentigo, melanosis, recurrent nevus, Reed nevus, seborrheic keratosis, and vascular lesion. There is also a small number of cases classified simply as `miscelaneous'.

Many works employ the Atlas (e.g. \citet{wadhawan11EMBC, wadhawan11ISBI, diLeo10HICSS, barata2015improving}), but comparability is seriously hindered due to  each work selecting unspecified subsets of the cases. As we will see, the ``difficult'' images are \textit{extremely} difficult, so it makes sense to remove them from exploratory experiments. Some diagnostic classes (melanosis, recurrent nevus), or lesions with certain clinical characteristics (e.g., lesions on the palm of the hands, sole of the feet, or the genital mucosa) are notoriously problematic, even for humans. Hair, rulers, bandages or other artifacts may hinder the analysis --- thus images containing those artifacts are often excluded. The issue is not the selection itself, but the uncontrolled variability brought by an \textit{unknown} selection.

In our experiments we employed three subsets: LM, LM+, and LMH. In all subsets, we excluded the images with large occlusions by rulers. Some images contained a black ``frame'' around the picture, which we removed, cropping by hand. We kept both types of images of lesions: clinical and dermoscopic, but excluded clinical images that showed large parts of the body instead of lesion close-ups. In LM and LMH we also removed the images with hair, kept in LM+. LM and LM+ contained only images of low- and medium-difficulty diagnosis, while LMH contained also high-difficulty images. In all subsets, all subtypes of melanoma were considered positive, and all other cases were considered negative. LM ended with 918 images (250 positive), LM+ with 1052 images (267 positive), and LMH with 1271 (364 positive). As everything else, those choices are far from obvious --- we will further discuss the complex issue of selecting subsets and annotations in data in the final section.

The Atlas is far from perfect. It is publicly available --- for a fee\footnote{We share hints for acquisition at our code repository.}. The fact the Atlas is intended for human training, and not for machine learning, is both a strength and a weakness. On one hand, we can count on a large diversity of representative cases, but also on difficult cases needed for differential diagnosis. We can also count on reliability: all but the simplest cases had the diagnosis confirmed by excision and subsequent histopathologic exam. On the other hand, the Atlas is not meant for automatic usage. Its metadata (diagnostic, clinical, etc.) are in a cumbersome format, protected by layers of encryption. We were lucky to get the help of \citet{abbas12Skin}, who shared their own machine-readable version with us, otherwise the alternative would be laborious manual transcription and image matching. 

\subsection{Validation Protocols and Metrics}

The main validation protocol was 10-fold cross validation, motivated by the desire to approximate the pipeline of \citet{wadhawan11ISBI} as closely as possible.

The folds were divided by \textit{case}, to avoid the subtle train--test contamination when a single case splits over the two sets. We divided the 10 folds ``semi-randomly'' --- trying to balance them in terms of positive/negative ratio, types of lesion present in the negative class, and difficulty level --- an effort to control the sources of variability.

For all experiments, we measured the AUC, since it offers more global information than of a single Sensitivity-Specificity point. In that spirit, we also traced the entire ROC curves.

    \section{Results}
\label{sec:results}
Table~\ref{tab:results} shows the average AUC over the 10 folds. Without surprise, there is a progression between [Simple Texture Features + Classical BoVW] $\rightarrow$ [Discriminating Invariant Features + Advanced BoVW] $\rightarrow$ [Deep Learning + Transfer Learning], reflecting the recent history of Computer Vision. The bet on Deep Learning winning over Advanced BoVW was, perhaps, not so obvious, but shows the efficacy of transfer learning in making the former feasible for small medical datasets.

\begin{table}[tb]
    \centering
    \caption{Main results}
    \label{tab:results}
    \begin{tabular}{ccccc}
    \toprule
     \multirow{4}{*}{\textbf{Group}}  &  \multirow{4}{*}{\textbf{Hair?}} & \multicolumn{3}{c}{\textbf{Average AUC over 10 folds (\%)}}  \\ \cmidrule{3-5} 
     & & \multirow{2}{*}{\textbf{Baseline}}        & \multirow{2}{*}{\textbf{BossaNova}}        & \textbf{Deep+} \\ 
     & &       &                                            & \textbf{Transfer} \\ \midrule
    \textbf{LM}  & \textbf{no}  & 80.6 & 83.6 & 89.2 \\ 
    \textbf{LM+} & \textbf{yes} & 81.2 & 84.6 & 89.3 \\ 
    \textbf{LMH} & \textbf{no}  & 76.4 & 77.3 & 79.0 \\ \bottomrule
    \end{tabular}
\end{table}

Those results are particularly interesting in view of the relative economy of the advanced models. While the baseline used  full-size images, BoVW and DNNs employed considerably reduced ones. The baseline has a complex segmentation procedure, followed by a complete BoVW pipeline. The BossaNova pipeline, while arguably implementing a more complex BoVW pipeline, forgoes segmentation completely. The Deep Learning model also forgoes segmentation, takes a \emph{completely pre-trained} feature extractor, and after some simple normalization, moves directly to SVM. That simplicity is revealed in run times: the two BoVW pipelines take many days to run from end to end, while the Deep+Transfer pipeline completes an experiment in a matter of \emph{minutes}.

\begin{figure}[bt]
    \hspace{-0.3cm}\includegraphics[width=1.1\columnwidth]{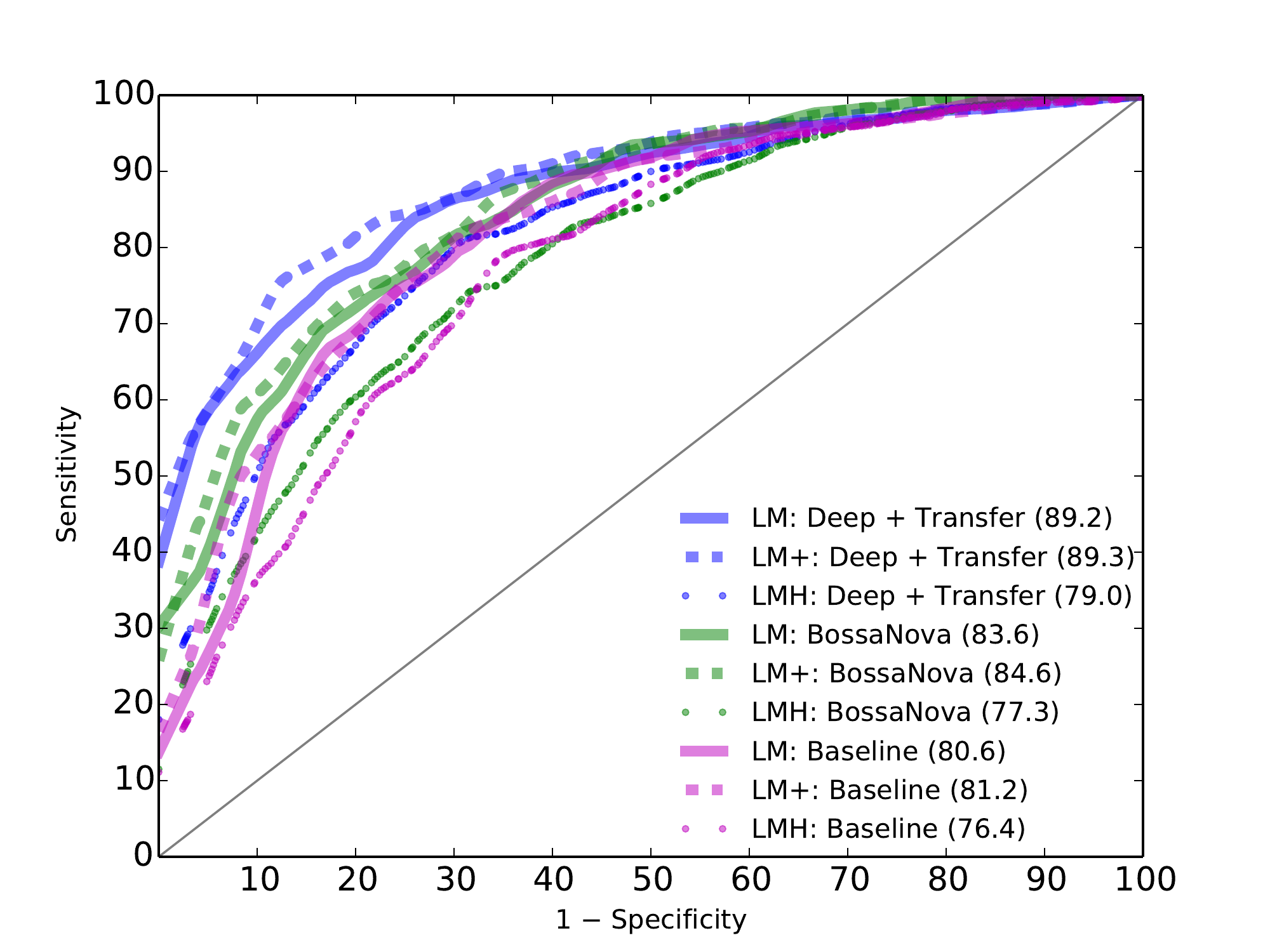}
	\caption{The ROC curves corresponding to the results shown in Table~\ref{tab:results}.}
	\label{fig:rocs}
\end{figure}

A more detailed view of the results appears in Figure~\ref{fig:rocs}, showing the full ROC curves, each averaged over the 10 folds. Throughout the curves, DNN dominates BossaNova, which dominates the baseline.

\begin{table*}[t!]
    \centering
    \caption{Results stratified by diagnosis difficulty of test images: Low, Medium, or High. The difficulty is subjectively established by humans when training doctors for diagnosis.}
    \label{tab:estratificado}
    \begin{tabular}{ccccccccccc}
    \toprule
    \multirow{4}{*}{\textbf{Group}} & \multirow{4}{*}{\textbf{Hair?}} & \multicolumn{9}{c}{\textbf{Average AUC over 10 folds(\%)}}                                                                                          \\ \cmidrule{3-11} 
                                   &                                 & \multicolumn{3}{c}{\textbf{Baseline}} & \multicolumn{3}{c}{\textbf{BossaNova}} & \multicolumn{3}{c}{\textbf{Deep+Transfer}} \\ \cmidrule{3-11} 
                                   &                                 & \textbf{L}   & \textbf{M}  & \textbf{H}  & \textbf{L}    & \textbf{M}  & \textbf{H}  & \textbf{L}   & \textbf{M}   & \textbf{H}   \\ \midrule
    \textbf{LM}                    & \textbf{no}                     & 85.3         & 72.1        & --          & 87.1      & 78.4       & --          & 93.4        & 81.7        & --            \\ \midrule
    \textbf{LM+}                   & \textbf{yes}                    & 85.8         & 75.3        & --          & 87.9      & 78.9       & --          & 93.1        & 82.7        & --            \\ \midrule
    \textbf{LMH}                   & \textbf{no}                     & 84.3         & 79.4        & 57.9        & 85.8      & 82.0       & 57.6        & 91.7        & 80.5        & 50.0         \\ \bottomrule
    \end{tabular}
\end{table*}

The presence of hair, constant worry of classical techniques, turns out to be a non-problem. None of the techniques suffered appreciably in the transition from LM to LM+. Actually, the results slightly improved --- due to LM+ being $\sim15\%$ larger. All techniques preferred a larger training set with hair than a smaller one without hair.

The scenario is much different for the transition to LMH. Even with a dataset $\sim38\%$ larger than LH, the results are much worse. Those results are better appreciated in Table~\ref{tab:estratificado}, which splits the results of Table~\ref{tab:results} according to the difficulty of the test image. For all models, the performance on difficulty images is a little better than random guess --- when the Atlas says ``difficult to diagnose'', it clearly means it! 

Try as we might, we simply could not reach anywhere close to the AUC of 91.1\% reported by \citet{wadhawan11ISBI}. The most probable cause is that we could not quite match their data and label selection, and our protocol ended much harder than theirs. Of course, there is always the chance that our implementation, having to fill so many missing details, differs from the original (see Section~\ref{sec:baseline}). Considering that theirs is one of the best described pipelines in the literature, those results reveal that in our community, works are effectively irreproducible.
\\

\noindent\textbf{Conclusions.} Our review identified two main challenges in the literature of melanoma screening: the use of outdated Computer Vision models, and reproducibility issues. We argue both claims are apparent in our results: given the same experimental setup, two advanced Computer Vision methods perform much better than a classical one, even though forgoing segmentation. Our best efforts to reproduce the experimental setup of the best described work in the literature fail to reach the numbers they report. We will discuss the implications of those findings in the next section.

    \section{Discussion and Call for Action}
\label{sec:conclusions}

\subsection{Reproducibility} 

Let us start by stating explicitly what this is article is \emph{not}. 

\emph{It is not a condemnation of an author, a work, or a group.}  A broad reimplementation campaign would be simply unfeasible: the reimplementation effort for \textit{one} baseline took us \textit{months}, and when we could not reproduce their numbers, we spent more than a year trying to debug code, data, and procedures. We were unfortunately forced to single out one article. But we firmly believe that, had we chosen another work, the challenges to reproduce it would be the same or worse. The lack of cooperation we faced seems --- and is --- unlaudable, but it was perfectly typical of our contacts with the community. Let the reader who never ignored a colleague's nagging e-mail cast the first stone.

\emph{It is not a hypocritical self-proclamation of virtue}. Although we are fighting to improve, reproducing our works can seldom happen without help. Old works, for which we lost custody of code or data, probably must be reimplemented from scratch. This article is not our attempt to take some kind of moral higher ground.

What is our proposal, then? As we mentioned, reproducibility is a thorny issue, plaguing all Science~\citep{allison2016reproducibility}, but in particular Computer Science~\citep{peng2011science,sandve2013plos}. We agree with \citet{allison2016reproducibility} that the process of scientific post-publication review, correction, or eventual retraction needs improvement. 

First of all, results must be verifiable, so that eventual mistakes be findable, i.e., papers must be reproducible. \citet{sandve2013plos} proposes ten relatively simple, and cost-effective ways to improve reproducibility in Computer Sciences. It is superfluous to repeat them here --- we invite the reader to study the open-access original --- but we highlight rules three (``archive exact versions of all external programs used''), four (``version control all custom scripts''), and ten (``provide public access to scripts, runs, and results''). 

We also need a better way to deal with honest mistakes. We do not advocate a culture of sloppiness, but when the only options to authors are standing behind results, or issuing a full retraction, it is difficult to create a culture of self-correction. \citet{allison2016reproducibility} mention a publisher who charges the authors USD 10K for a retraction. Those disproportionate penalties help neither authors nor readers. Given the extreme complexity of current processing pipelines, few --- if any --- authors in Computer Science will spend an entire career without mistakes. Some errors --- while needing correction, are made in good faith, and do not invalidate the essence of an article. We need publishing culture and practices reflecting that. Reclaiming a quote by Allison et al.,``robust science needs robust corrections. It is time to make the process less onerous''.

We attribute part of the problem to the disproportionate role of conferences in Computer Science scholarship, one of the few fields where proceedings are archival material. There is an ongoing discussion on the consequences of that culture~\citep{wainer2013happens}. However, a 4-page limitation is no longer an excuse to irreproducible results in a world with DOIs, Webs, public source code repositories, and preprint archives.

Commercial datasets, like the one we use, pose special challenges in terms of reproducibility, due to copyright and the risk of legal action. In our case, we decided on a compromise, disclosing information that should be considered ``fair use'', enough for our scientific peers to reproduce, or improve our results, but not enough to hurt EDRA/DS Medical commercially at their intended audience.

\subsection{Experimental Validity}

Any machine learning result in less than a thousand samples is suspicious, to say the least. Models quickly learn the biases of small datasets, presenting fantastic results that do not generalize later.  

Train--test contamination is a major concern. Obvious contamination (e.g., the same file on the two sets) seldom occurs, but more subtle forms may happen: e.g., computing some seemingly innocent statistic (an average, a standard deviation) over the entire dataset and employing it on the model. A more insidious form of contamination is performing too many cycles of meta-parameter adjustment over a small dataset, where the \textit{human operator} employs results over the test set to optimize the model. To vaccinate against contamination, we try to visualize our test sets at the far, far future, and thus think of any form of information from it as utterly inaccessible.

Ideally, researchers should adopt cross-dataset protocols, with two datasets similar in essential properties, but different in the nuisance factors the model is intended to resist: operator, equipment, time of the year, institution, etc. Researchers should do all model development and (meta-)optimization on one dataset, and use the other very sparingly, to stress-test the actual generalization ability of models, at the end of research cycles no shorter than a few months. We are, however, the first to recognize that having \emph{one} good dataset is already a luxury in medical research. Thus, having two good sets and keeping one strictly on the fridge requires extraordinary discipline. 

In that sense, public annual competitions like the one just organized by ISIC (end of Section~\ref{sec:problems}) are very helpful to the community, fulfilling the role of the ``frozen'', seldom used, test dataset. To play that role, however, competitions must withhold the test labels even after the results are announced, or use fresh test images each year.

Setting the positive/negative labels for automated melanoma research is far from obvious, and at the very least, requires careful documentation. For example: what to do with basal cell carcinoma? The answer depends on the aims of the work: patient referrability? Malignant vs. benign lesions? Differential diagnosis? In the latter case, would it be preferable to target three or more classes instead of simply positive/negative? What should we do with benign lesions known to mimic melanoma? Include them to attempt a crisper differential diagnosis? Or exclude them considering that those cases should be referred to the human doctor? Those are important questions for which there is no definitive answer, but which must be addressed as automated screening moves from theoretical possibility to practical solution. 

\subsection{Future Research}

We believe that past models (global descriptors, BoVWs) can no longer compete with DNNs, even in the context of small-dataset medical applications. If some model can beat DNNs, it is a model from the future, not from the past. In our group, our current research efforts on automated diagnosis rely on DNNs.

Our own future works intend to explore more in-depth the issue of dataset organization for experimental validity. We are widening the communication between the Computer Science and the Medical Science partners of this work, in order to better answer those questions.

Many challenges lie ahead of automated melanoma screening as a practical tool, beyond the issues of machine learning: for example, usability and privacy. As computer vision models improve fast, we expect the community will soon move to those (harder) issues. We are eager to reach that exciting turning point.

    \section*{Acknowledgements}
    This research was partially supported by FAPESP, CAPES, and CNPq. We thank Amazon Web Services, Microsoft Azure, and CENAPAD (Project 533) for the infrastructure used on the experiments. We are very grateful to Prof. Dr. M. Emre Celebi for kindly providing the machine-readable matadata of The Interactive Atlas of Dermoscopy. We are grateful to the student Tiago Ferreira for the first reimplementation of \citet{wadhawan11ISBI}'s work, from which we based the version used in this paper.

    \balance
    \bibliographystyle{IEEEtranN}
    \bibliography{references}
    
\end{document}